\documentclass[letterpaper]{article}
\usepackage{interspeech2010,amssymb,amsmath,epsfig}
\usepackage[safe]{tipa}
\usepackage{tipx,nicefrac,url}
\usepackage{fixltx2e} 
\def\reg{{\rm\ooalign{\hfil
     \raise.07ex\hbox{\scriptsize R}\hfil\crcr\mathhexbox20D}}}

\usepackage{listings}
\usepackage{color}
\title{Stitched Panoramas from Toy Airborne Video Cameras}

\makeatletter
\def\name#1{\gdef\@name{#1\\}}
\makeatother
\name{\em Camille Goudeseune}
\address{Beckman Institute, University of Illinois at Urbana-Champaign (UIUC) \\
{\small \tt cog@illinois.edu}}

\begin{document}
\maketitle
\begin{abstract}

Effective panoramic photographs are taken from vantage points that are high.
High vantage points have recently become easier to reach as the
cost of quadrotor helicopters has dropped to nearly disposable levels.\footnote{
Disposal is trickier than it sounds.  I have reclaimed such aircraft undamaged after
extended periods on roofs, in trees, among cattle, and under fast-moving cars.}
Although cameras carried by such aircraft weigh only a few grams,
their low-quality video can be converted into panoramas of high quality and high resolution.
Also, the small size of these aircraft vastly reduces the risks inherent to flight.

\end{abstract}
\section{Introduction}

High-quality panoramic photographs can now be acquired from aircraft under 100 grams.
This is desirable because these ``toys'' pose a far smaller risk than aircraft
carrying a camera that itself weighs more than 100 g.
(Quality cameras are so heavy because of their glass lenses.
This is unlikely to change soon.)
The risk reduction can be quantified by estimating their reduced
gravitational potential energy ($0.05 \times$ mass, $0.3 \times$ height: $0.015 \times$ net),
kinetic energy (mass as before, $0.4 \times$ airspeed: $0.008 \times$),
rotor kinetic energy (about $0.05 \times$),
and battery energy in mWh ($0.05 \times$).
Financial risk due to aircraft damage or loss is also reduced about twentyfold.
Photography from places too confined or too risky for larger aircraft becomes possible.
Also, an aircraft small and light enough to always keep with you encourages
impromptu photography: these days, SLR cameras take far fewer photos
than mobile phones do.

Capturing video from sub-100 g aircraft is common~\cite{microquad},
but no reports have been published about capturing still photographs.
This document's novel contribution is a complete set of techniques for acquiring high-quality panoramas from these aircraft:
how to
maneuver effectively,
cope with wind,
extract still frames from a video recording,
robustly and automatically cull frames to avoid motion parallax and motion blur,
suppress artifacts due to the camera's poor quality,
and record simultaneously from multiple cameras.
These techniques are all simple and inexpensive, as they should be for a toy.

\begin{figure}
\centerline{\epsfig{figure=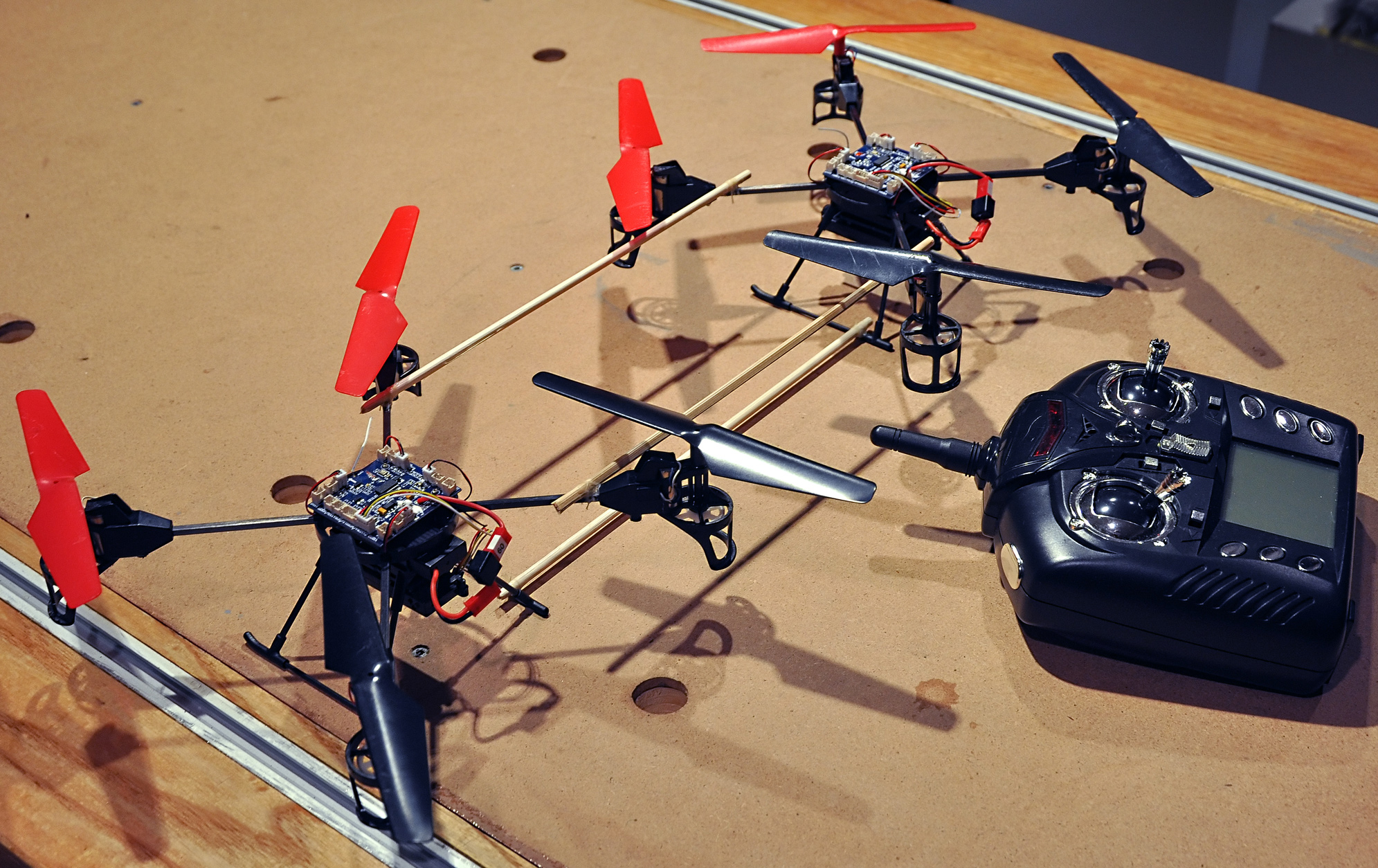, width=\linewidth}}
\caption{Two videocamera-equipped quadcopters, with a shared radio-control transmitter.} 
\label{fig:workbench}
\end{figure}

\subsection{Quadcopters}

From 1990 to 2000, electric power for radio-controlled aircraft
developed from a curiosity to a commonplace, as batteries and motors
improved to match the sheer power of piston engines.
Erasing that performance deficit left the electric drivetrain with only
advantages, notably reliability, less vibration,
and mechanical simplicity---often only one moving part.

During the next decade, electric and electronic technology continued to improve,
while consumer preference for mechanical simplicity remained high.
This technological progress then produced another commonplace: the quadrotor helicopter, or quadcopter.
Because differential thrust controlled pitch, roll, and yaw, neither
servomotors nor swashplates were needed, leaving the entire aircraft
with only four moving parts.
Accelerometers and gyroscopes made flight easy to learn.
Pushing all of the aircraft's complexity into software
made it inexpensive to manufacture, maintenance-free,
easy to repair, and crash-resistant---if only because it weighed less
than a gerbil.  The price of camera-equipped quadcopters,
such as those in fig.~\ref{fig:workbench}, has fallen to USD 45~\cite{myrcmart}.
(The larger quadcopters that record sporting events are quite the opposite:
many moving parts, fussy maintenance, high fragility, and a price in the thousands of dollars.)

\begin{figure*}[htbp] 
\centerline{\epsfig{figure=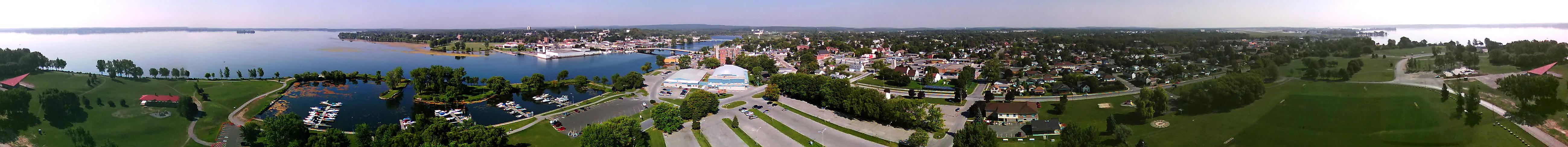, width=\linewidth}}
\caption{Full $360^{\circ}$ panorama. Trenton, Ontario, 2013-08-19.}
\label{fig:trenton}
\end{figure*}

Sub-100 g aircraft are inconspicuous and quiet: using one I photographed
a family wedding's outdoor reception, without anybody noticing.
Quadcopters much lighter than 100 g are now available, but
would have been uncontrollable in the gusty 10 knot winds that day
(this 75 g one just managed).
Although flying animals much smaller than that shrug off such winds,
a hummingbird's performance won't soon be matched by consumer goods.
A flying weight near 100 g will likely remain optimal for a few years.
This small size also permits unplanned opportunities to be exploited,
such as fig.~\ref{fig:trenton}, captured in midmorning while waiting
ten minutes for the beer store to open.  As the saying goes, the best
camera is the one you have with you.

\section{Converting Video to a Panoramic Photo} 
\label{sec:ffmpeg}

The inexpensive videocamera commonly used for stealth or light weight
has no official name.  Vendors call it a keychain camera;
hobbyists call it an ``808''~\cite{eight-o-eight}.  Its attributes
have changed monthly for some years, but are roughly:
weight 8~g, pixel resolution $640\times480$ to $1280\times800$,
microSD card storage,
30 or 60 frames per second,
fixed focus,
and depth of field 10 cm--$\infty$.
Its 2~mm diameter lens performs poorly in low light.

The camera saves a video file in motion JPEG format, which is just a
soundtrack combined with individual JPEG images~\cite{mjpeg}.
Because this format does not exploit inter-frame redundancy,
it produces files 3 to 10 times larger than those made with the modern H.264 codec.
This size is acceptable, though, because it does not
constrain recording---an 8 GB card easily stores a dozen 5-minute flights.
In fact, were the camera's CPU advanced enough to compress video better,
its increased power consumption would deplete the battery faster,
paradoxically decreasing the duration of both a flight and its recording.

Individual frames from the video file can be extracted
with the open-source software FFmpeg~\cite{ffmpeg}:

{\tt ffmpeg -i in.avi -vsync 0 -vcodec png -f image2 \%04d.png}

This command produces images named {\tt 0001.png}, {\tt 0002.png}, ..., {\tt 1138.png}.\footnote{
Alternatively, the original JPEG frames can be very quickly extracted:
{\tt ffmpeg -i in.avi -vsync 0 -vcodec copy -f image2 \%04d.jpg}.  (The option
{\tt -vcodec jpg} should be avoided, because it transcodes and
further degrades each frame instead of just extracting it.)
This shortcut is convenient for video good enough to need no improvement
with the tools listed in sections \ref{sec:artifacts} and \ref{sec:blur}.}

Dropped or missing frames occur with some camera--card combinations,
or when the camera's CPU is momentarily too slow.
Na\"{\i}ve extraction of frames ``reconstructs'' these missing frames by
repeatedly duplicating the previous frame;  this duplication would
slow down image stitching.\footnote{
Proper interpolation, which analyzes frame-to-frame motion,
has been implemented for some keychain cameras~\cite{repair808},
but this interpolation improves only video, not stitched panoramas.}
Many of these consecutive duplicate frames are removed with FFmpeg's option {\tt -vsync 0}.
Removing \emph{all} duplicate frames requires
a duplicate-file finder, such as the Linux command {\tt fdupes --delete --noprompt *.png}.
Because these finders use file size as a quick
first test for duplication, they are much slower with formats that
give every frame the same file size, such as {\tt.bmp} and {\tt.ppm}.
The {\tt.png} format does not suffer from this.

These image files are sent to an automatic image stitcher,
such as the free programs AutoStitch~\cite{autostitch,autostitch2} and
Image Composite Editor~\cite{ICE}.  The stitcher then produces a
single panoramic image (figs. \ref{fig:trenton} and \ref{fig:octojenny}).

\begin{figure}
\centerline{\epsfig{figure=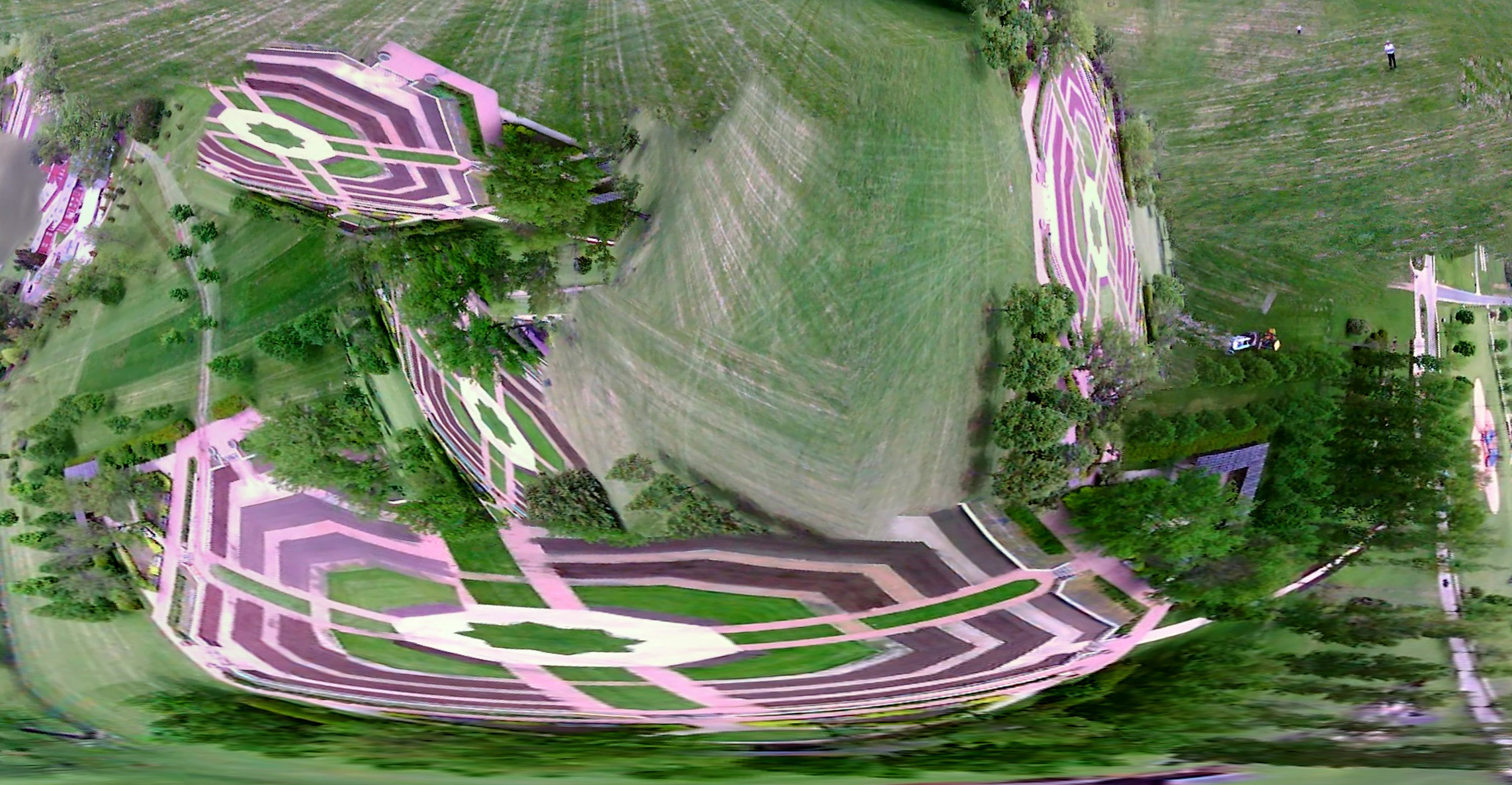, width=\linewidth}}
\caption{Mis-stitching due to camera movement.  UIUC Arboretum, Urbana, Illinois, 2013-05-16.}
\label{fig:dali}
\end{figure}

\section{Flight Path}

An image stitcher must assume that the photos it is given were
taken from a single viewpoint.  Because a quadcopter is
hardly a stationary tripod, stitching the video recording of an entire flight
spectacularly violates this assumption (fig.~\ref{fig:dali}).
For a coherent panorama, only a subinterval of the recording should be stitched.

A convenient way to record a stitchable subinterval is to yaw (pirouette)
the quadcopter while hovering.  Some drifting is
tolerable if the subject is not very nearby,
and if the pirouette is less than a complete circle.
Stitching is improved when frames have more overlap, which happens with slower yaw.
The slowest practical yaw for a 100 g quadcopter is
about 0.4~rad/s (16~s for a full pirouette);
this is rarely slow enough to introduce other problems like ghosting~\cite{chenSICE,ghost}.

\begin{figure*}[htbp]
\centerline{\epsfig{figure=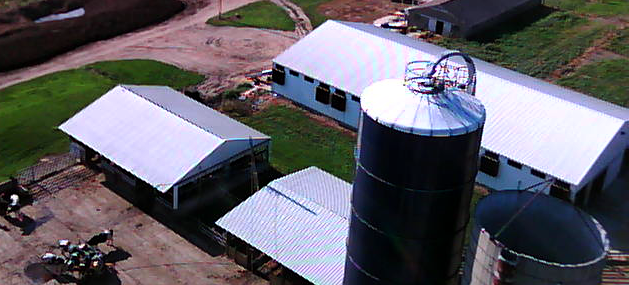, width=\linewidth}}
\caption{Different magenta-cyan moir\'e patterns on three identically corrugated roofs.  The roofs differ only in their distance from the camera. UIUC Dairy Cattle Research Unit, 2013-08-01.}
\label{fig:moire}
\end{figure*}

\subsection{Choosing a Stitchable Subinterval}

After landing, the video is viewed on a computer to find an interval where
the desired subject is visible.  To maximize the panorama's coverage,
the interval's endpoints are extended, with two constraints (often identical):
exclusion of non-yaw flight and exclusion of different viewpoints of the subject.

These constraints generally restrict the interval to a single monotonic yaw maneuver.
One might think that several back-and-forth pans cover the subject
more widely and give more information to the stitcher;
but in practice each pan is from a slightly different location,
introducing seams like the one in the fence in fig.~\ref{fig:octojenny}.

The position of the interval's endpoints
and the duration of the entire video determine the endpoints
as a fraction of the video's duration.  These fractions then approximate
the filenames.  For example, consider a
video lasting 100~s, with a stitched interval starting at 50 s and ending at 60 s,
and filenames {\tt 0000.png} to {\tt 3000.png}.
The interval's filenames will be approximately {\tt 1500.png} to {\tt 1600.png}.
Because of duplicated frames, the numbers 1500 and 1600 are
only approximate; manual verification is needed.

\subsection{Video Downlink}

After a few flights, most pilots develop an intuition for what the
quadcopter's camera is seeing.  But if the quadcopter's height exceeds
that used to capture fig.~\ref{fig:trenton}, about 30~m, it becomes
almost too small to see, let alone aim its camera.  If this is
a concern, a live video downlink can be added.
But this expense is substantial compared to the stock quadcopter,
because many parts must be removed or replaced with lighter ones to
compensate for the extra payload~\cite{archilleselbow}.
Also, such first-person view (FPV) flight is riskier because of its
many single points of failure.  Even for a 100 g aircraft, never mind
a 5~kg one, the prudent FPV pilot keeps the aircraft near enough for line-of-sight control,
and asks an assistant ``spotter'' to maintain situational awareness.

\section{Suppressing Camera Artifacts}
\label{sec:artifacts}

A keychain camera's poor image quality may be evident in several ways:
varying brightness and color, rolling shutter, moir\'e bands, and JPEG compression blockiness.
Fortunately, these artifacts can be suppressed or even eliminated.

\subsection{Varying Brightness and Color}

Variations in brightness and color are due to the camera's automatic
exposure compensation and automatic white balance~\cite{ghost}.
When the view changes suddenly from, say, bright cumulus clouds to
tree-shaded terrain, the camera takes a second or two to correct its exposure.
Similarly, when a view of only grass suddenly tilts up to include sky,
it takes a second for the white-balanced grayish grass to become bright green.
Frames from such transitions may not be usable.

Reducing such variations requires slower aircraft rotation.
After flight it may be too late to correct the transitional frames
if color is out of gamut, or if shadows or highlights are clipped
(lost detail, in pure black shadows or pure white highlights).

\subsection{Rolling Shutter}

Rolling shutter is a motion artifact common to small cameras:
the image is captured one scanline at a time, instead of all at once.
In other words, different parts of the image correspond to different instants in time.
Therefore, moving the camera relative to the subject produces visible warp and skew.
(This can be demonstrated by waving one's hand in front of a photocopier's scanner as it slides along.)
As with varying brightness, slower aircraft rotation is the first cure.
Also, balancing the propellers with flecks of adhesive tape reduces
mechanical vibration, which causes what hobbyists call ``jello'' in video~\cite{graham}.

Unlike varying brightness, though, rolling shutter can also be suppressed after flying~\cite{baker,grundmann}.
Rolling shutter repair is included in commercial video software such as Adobe Premiere Pro and Adobe After Effects,
and in free video software such as the Deshaker~\cite{deshaker} plug-in for VirtualDub~\cite{virtualdub}.
However, these tools specialize in inter-frame smoothness, which panorama stitching does not need.
Worse, they may crop the image (which reduces the panorama's coverage),
or add a black border (which confuses the stitcher).
If the border's color can be made transparent, however,
commercial stitchers such as Adobe's Photomerge may succeed.
Better yet, Deshaker can fill the border with pixels from previous or successive frames,
or, when those are unavailable, with colors extrapolated from the current frame.

These tools require the individual frames to be re-encoded as a video file,
to give the detection-and-removal algorithm more material to work with:
several successive frames of the same subject, not just a single frame.

\subsection{Moir\'e}

The artifact called a moir\'e pattern consists of undesired bands of hue or brightness (fig.~\ref{fig:moire}),
seen in a subject with repetitive detail, such as stripes, that exceeds the camera's resolution.
(The pattern is due to foldover at the camera's Nyquist frequency.
Non-toy cameras suppress these patterns with anti-alias filters.)
If the stitcher tries to match these bands, which shift from frame
to frame as the camera moves, stitching quality is reduced.  This is
particularly so for stitchers that match image features by hue as well
as by brightness, because a camera sensor's Bayer filter mosiac easily
produces hue bands.

Avoiding moir\'e patterns requires such subjects to be either very distant,
or so close that each stripe is at least two pixels wide (for a keychain
camera, at most a few hundred stripes visible at once).

\subsection{JPEG Compression Blockiness}
\label{sec:unblock}

Some JPEG frames are compressed so strongly that a grid appears
at the boundary between $8\times8$ blocks of pixels.
As with moir\'e patterns, this noise varies from frame to frame,
slightly distracting the stitcher from finding
common elements across frames.  It also looks ugly in the final panorama.

This artifact is suppressed by the UnBlock algorithm~\cite{costella,unblock},
which smooths over the boundaries between blocks, but only aggressively
enough to reach the same distribution of discrepancies across the block
boundaries as is found in the block interiors (fig.~\ref{fig:unblock}).
This approach prevents worse artifacts from being introduced as a side effect.
The algorithm also needs no tuning.

\section{Kites}

\begin{figure}
\centerline{\epsfig{figure=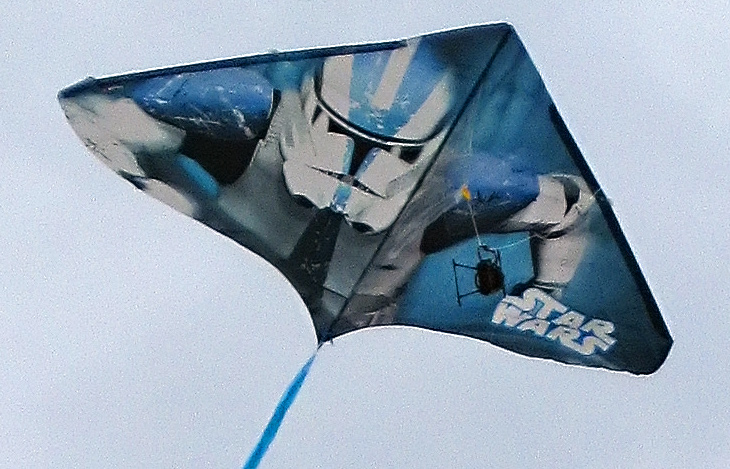, width=\linewidth}}
\caption{Kite hoisting a rotorless quadcopter-camera (a ``nullicopter''), while capturing fig.~\ref{fig:kiteblur}.}
\label{fig:kite}
\end{figure}

In winds too strong for a lightweight quadcopter, it can nevertheless be given a high
vantage point by hanging it from a toy delta-wing kite (span 1.3~m, cost USD~5).
Even with its four booms removed to prevent its propellers from getting tangled in the kite's tether,
it may still operate as a power source and remote control for the camera (fig.~\ref{fig:kite}).

\begin{figure}
\centerline{\epsfig{figure=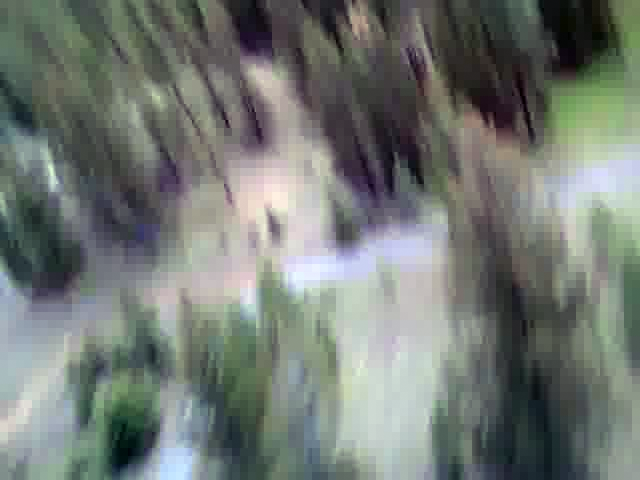, width=\linewidth}}
\caption{Strong motion blur from a kite-suspended camera.  Evergreens 5 to 15 m tall, Okanogan-Wenatchee National Forest, 2013-05-22.}
\label{fig:kiteblur}
\end{figure}

An elaborate Picavet suspension~\cite{picavet2,picavet} for the camera
is not in the spirit of cheap, simple hardware.
On the other hand, just dangling the camera from the tether
can cause so much camera shake that fewer than one frame in a
hundred is usable for stitching (fig.~\ref{fig:kiteblur}).  Happily, the shaking
can be dampened by hanging the camera from not one but two points on the tether,
at the bottom of a `V.' Then one frame in ten has acceptably low motion blur.

\subsection{Motion Blur}
\label{sec:blur}

Manual culling of frames blurred by camera motion is impractical.
To automate this, one can measure how blurred each frame is, and then sort the
frames by blurriness with a Schwartzian transform.
Blurriness can be measured simply and thus robustly
by re-saving the frame in JPEG format, with and without first applying a Gaussian blur.
The smaller the ratio of the sizes of the two resulting files,
the less difference the Gaussian blur made,
and thus the blurrier the original frame.
(A more elaborate method,
culling any frame that has few sharp edges compared to its neighboring frames~\cite{chenISIC},
fails in the presence of the duplicate frames mentioned in section~\ref{sec:ffmpeg}).

\begin{figure}
\centerline{\epsfig{figure=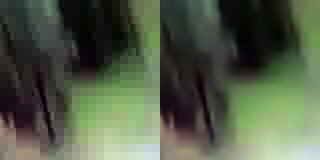, width=\linewidth}}
\caption{Detail ($160\times160$ pixels) from top right of fig.~\ref{fig:kiteblur}. Left: original.  Right: processed by the UnBlock algorithm.}
\label{fig:unblock}
\end{figure}

This algorithm is implemented by the Ruby script in listing~\ref{ruby}.
It uses the ImageMagick program {\tt convert} to read, blur, and save files.
Because the script's performance is strongly dominated by the blur computation,
downsampling precedes the blur to speed it up sixteenfold.
The downsampling also attenuates the sharp pixel-block boundaries
described in section~\ref{sec:unblock} (fig.~\ref{fig:unblock}, left).  This is desirable because these
sharp boundaries reduce how well the Gaussian blur approximates the
original motion blur---they hide the smooth motion blur behind artificial crisp edges.
Finally, each file is given a symbolic link from a new directory, so the new directory
contains filenames sorted by blurriness rather than by time, for convenient manual inspection.

\begin{lstlisting}[caption=Ruby script to sort frames by blurriness., label=ruby]
#!/usr/bin/env ruby
$src = "/my_dir/frames_from_video"
$dst = "/my_dir/frames_sorted_by_blur"
`rm -rf #$dst; mkdir -p #$dst`

$tmp = "/run/shm/tmp" # fast ramdisk
$a = "#$tmp/a.jpg"
$b = "#$tmp/b.jpg"
`mkdir -p #$tmp`

pairs = []
Dir.glob($src + "/*.jpg") {|filename|
  `convert #{filename} -resize 25%                  -quality 50 #$a`
  `convert #{filename} -resize 25% -gaussian-blur 4 -quality 50 #$b`
  blur =  File.size($b).to_f / File.size($a) rescue 0.0
  pairs << [filename,blur]
}

pairs.sort_by! {|filename,blur| blur}

pairs.each_with_index {|(oldname,blur),i|
  newname = ('%05d' % i) + ".jpg"
  `ln -s #{oldname} #$dst/#{newname}`
}
\end{lstlisting}

Of course, a Gaussian blur only approximates a motion blur.  But the
exact motion blur is a combination of axial rotation and panning, which is too
expensive to measure for this quick first pass that culls almost all of the frames.
Later passes can use advanced algorithms~\cite{deblur2,deblur}, which can
not only detect but even remove mild blur by estimating camera motion
from consecutive frames---although these again fail for duplicate frames.
This advanced deblurring can also improve non-kite video.

\section{Multiple Cameras}
\label{sec:octocopter}

\begin{figure}
\centerline{\epsfig{figure=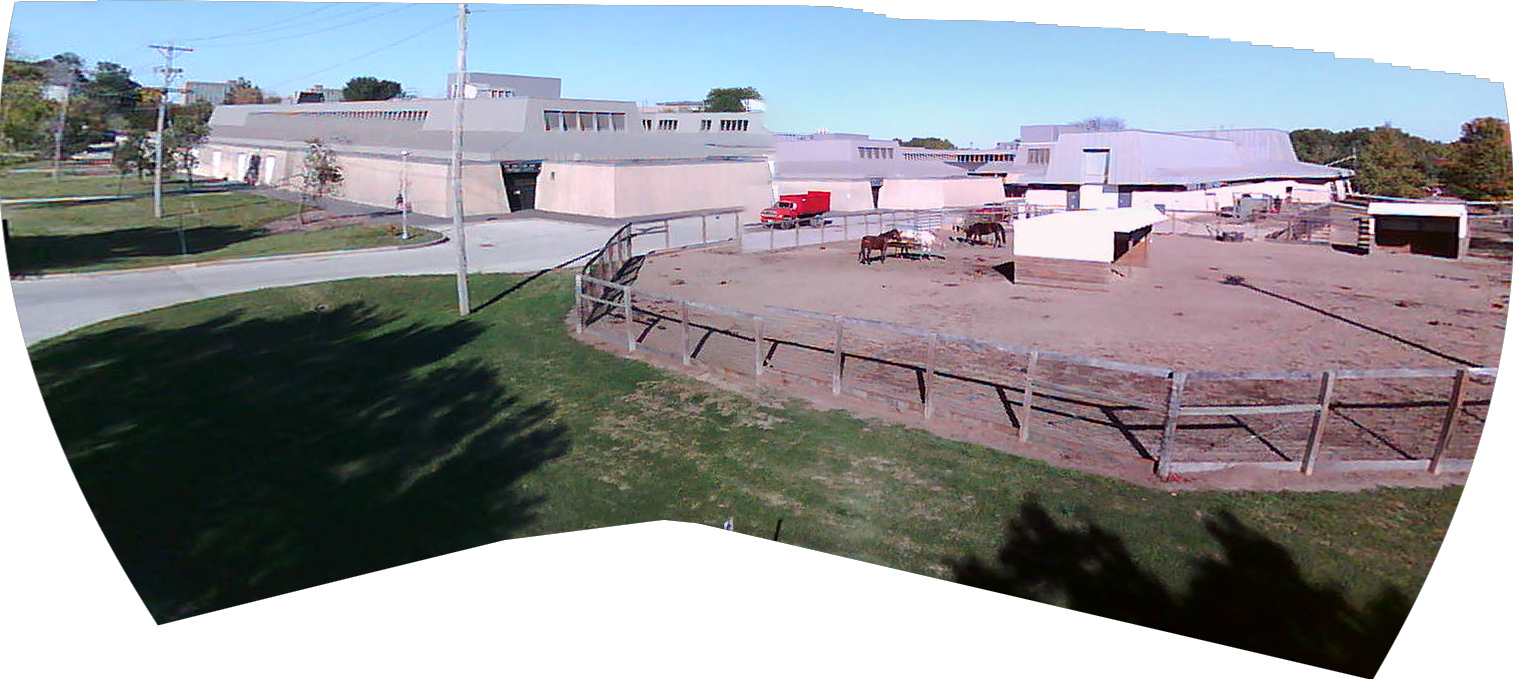, width=\linewidth}}
\centerline{\epsfig{figure=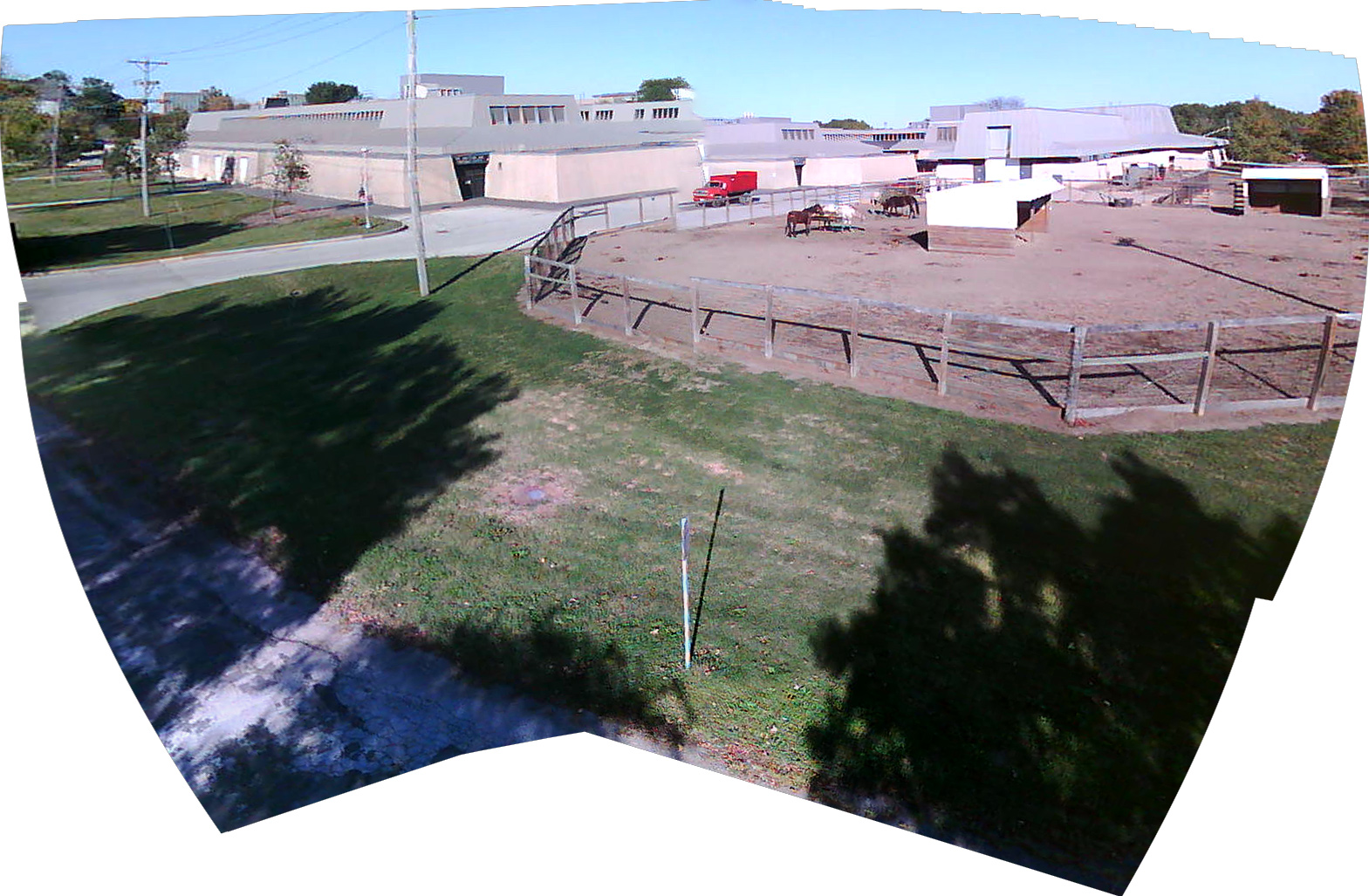, width=\linewidth}}
\caption{Top: panorama stitched from one camera's frames.  Bottom: second camera's frames added.  UIUC Large Animal Clinic, 2013-10-09.}
\label{fig:octojenny}
\end{figure}

A quadcopter may have enough thrust to carry more than one camera.
If each camera points in a slightly different direction,
the panorama gets more coverage (fig.~\ref{fig:octojenny}).
This has been proposed
for Parrot's AR.Drone quadcopter (400~g, USD~400)~\cite{chenISIC},
but no implementations to date have used sub-100 g aircraft.
More typical is DARPA's ARGUS-IS cluster of several hundred cameras~\cite{llnl}.

\begin{figure}
\centerline{\epsfig{figure=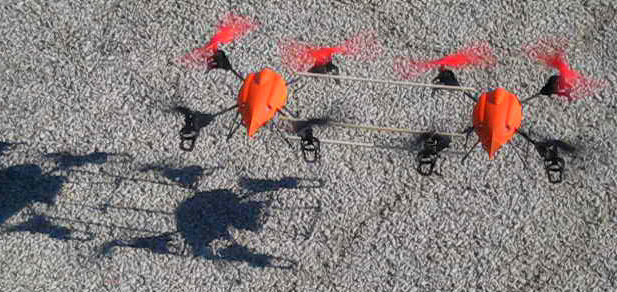, width=\linewidth}}
\caption{Two-camera octocopter, just before capturing fig.~\ref{fig:octojenny}.}
\label{fig:octocopter}
\end{figure}

If the quadcopter's maneuverability suffers with the extra payload of more cameras,
another novel solution is to laterally combine two or more quadcopters into an
octocopter (fig.~\ref{fig:octocopter}), dodecacopter, or hexadecacopter.\footnote{
Owning several quadcopters is not unusual: it is an inexpensive way to buy spare parts,
because a significant part of a quadcopter's mail-order cost is shipping.}
Bamboo skewers make good struts, being cheap, lighter than even a keychain camera, and almost as stiff as carbon fiber.
(The transmitter in fig.~\ref{fig:workbench} is unaware that it is controlling more than one quadcopter.)
The composite aircraft is slightly less maneuverable because the
stabilizers in each quadcopter fight each other, and
because roll authority is reduced.  But the more important
controls---pitch, yaw, and overall thrust---have no reduced authority.
As with multiple cameras on one quadcopter, each camera points at a different angle.

\section{Future Work}

Multiple cameras can record stereoscopic video, especially when
mounted far apart (large interpupillary distance) on an octocopter.
Sound recorded with each camera's rudimentary microphone helps to
synchronize the individual recordings.

Stereoscopic stitched panoramas can be made with only one camera,
recording two partial pirouettes from nearby locations
(half pirouette left, scoot forward a few seconds,
then half pirouette right).

An objective measure for the quality of image processing pipelines could
be constructed.  The challenge, for both synthetic imagery and
hundred-frame excerpts from actual flights,
would be the continually changing attributes of keychain cameras.

\section{Conclusion}

High-quality panoramic photos can be captured with a videocamera-equipped
quadcopter of startlingly small size, low cost, and low quality, thanks to
multiple stages of software post-processing.  These stages can be applied
to whichever aspects of a particular panorama need improving.

Basic piloting skill is needed, but the more fundamental skill is
choosing where to fly and when not to fly.  Even without these skills,
though, loss of flight control presents a hazard hardly greater than that of a stray Frisbee.
The same cannot be said of an aircraft powerful enough to carry
a 100~g camera.\footnote{
However, a secondary hazard can be posed by flying a 100 g quadcopter in public.
Because non-aeromodelers often lump together the risks of \emph{all} aircraft too small
to actually sit in,
pilots should avoid misleading bystanders
into thinking that a larger aircraft in that situation would pose no greater hazard.
}

For help in preparing this manuscript, I thank
Kevyn Collins-Thompson,
James A. Crowell,
Audrey Fisher,
Farouk Gaffoor,
David Gee,
Michel Goudeseune,
and
David Schilling.

\bibliographystyle{IEEEtran}

\end{document}